\documentclass[11pt,twoside]{article} 

\usepackage[utf8]{inputenc}
\usepackage{graphicx}
\usepackage{amsmath, amssymb, bm}
\usepackage{hyperref}
\usepackage{natbib}
\usepackage{fancyhdr}
\usepackage{geometry}
\usepackage{titlesec}
\usepackage{lipsum}
\usepackage{orcidlink} %
\usepackage{booktabs}
\usepackage{placeins}
\usepackage{float} 
\usepackage{placeins} 

\setcitestyle{authoryear}
\hypersetup{
	colorlinks=true,
	linkcolor=blue,
	filecolor=blue,
	citecolor=blue,
	urlcolor=black
}

\fancypagestyle{plain}{  
	\fancyhf{}
	
}

\title{
	\vspace{1em}
	\hrule height 1.5pt
	\vspace{0.4em}
	Structured Kernel Regression VAE: A Computationally Efficient Surrogate for GP-VAEs in ICA
	\vspace{0.8em}
	\hrule height 1.5pt
	\vspace{1em}
}

\author{%
\begin{minipage}[t]{.48\textwidth}\centering\small
  \textbf{Yuan-Hao Wei\orcidlink{0000-0001-9439-0780}}\\
  Hong Kong Polytechnic University \\
  \texttt{Yuan-Hao.Wei@outlook.com}
\end{minipage}\hfill
\begin{minipage}[t]{.48\textwidth}\centering\small
  \textbf{Fu-Hao Deng\orcidlink{0000-0001-8034-6949}}\\
  Hong Kong Polytechnic University \\
  \texttt{fuhao.deng@connect.polyu.hk}
\end{minipage}\\[3em] 
\begin{minipage}[t]{.48\textwidth}\centering\small
  \textbf{Lin-Yong Cui\orcidlink{0009-0005-9142-1728}}\\
  Hong Kong Polytechnic University \\
  \texttt{linyong.cui@polyu.edu.hk}
\end{minipage}\hfill
\begin{minipage}[t]{.48\textwidth}\centering\small
  \textbf{Yan-Jie Sun\orcidlink{0000-0002-7967-6382}}\\
  Hong Kong Polytechnic University \\
  \texttt{Yanjie.Sun@connect.polyu.hk}
\end{minipage}
}

\date{}

\begin{document}
\maketitle
    \thispagestyle{plain} 	
    \begin{abstract}
        The interpretability of generative models is considered a key factor in demonstrating their effectiveness and controllability. The generated data are believed to be determined by latent variables that are not directly observable. Therefore, disentangling, decoupling, decomposing, causal inference, or performing Independent Component Analysis (ICA) in the latent variable space helps uncover the independent factors that influence the attributes or features affecting the generated outputs, thereby enhancing the interpretability of generative models. As a generative model, Variational Autoencoders (VAEs) combine with variational Bayesian inference algorithms. Using VAEs, the inverse process of ICA can be equivalently framed as a variational inference process. In some studies, Gaussian processes (GPs) have been introduced as priors for each dimension of latent variables in VAEs, structuring and separating each dimension from temporal or spatial perspectives, and encouraging different dimensions to control various attributes of the generated data. However, GPs impose a significant computational burden, resulting in substantial resource consumption when handling large datasets. Essentially, GPs model different temporal or spatial structures through various kernel functions. Structuring the priors of latent variables via kernel functions—so that different kernel functions model the correlations among sequence points within different latent dimensions—is at the core of achieving disentanglement in VAEs. The proposed Structured Kernel Regression VAE (SKR-VAE) leverages this core idea in a more efficient way, avoiding the costly kernel matrix inversion required in GPs. This research demonstrates that, while maintaining ICA performance, SKR-VAE achieves greater computational efficiency and significantly reduced computational burden compared to GP-VAE.
    \end{abstract}
	
    \section{Introduction}
        The interpretability of generative models is considered a key factor in demonstrating their effectiveness and controllability. In general, it is believed that the distinct characteristics exhibited by different observed data are determined by latent variables, establishing a causal relationship between latent variables and observed data. Since the observed data are known, the construction of a generative model hinges on inferring the latent variables, the underlying “causes” that lead to the observed “effects.” Therefore, the inference process in generative models is essentially analogous to the inversion process in Independent Component Analysis (ICA) problems. In the architecture of Variational Autoencoders (VAEs)(\cite{kingma2013auto}; \cite{rezende2014stochastic}; \cite{kingma2019introduction}), the encoding process essentially infers the latent variables, while the decoder ensures that these latent variables can reconstruct the observed data. In other words, the encoder-decoder architecture of VAEs forms a closed loop of 'effect-cause-effect'. To enhance the interpretability of generative models, we strive for the dimensions of latent variables to be as independent as possible. This highlights the independent factors underlying the observed data and elucidates the data generation mechanism. Under ideal conditions, the latent variables of the observed data are accurately inferred, and the mapping process from these latent variables to the observed data is clearly revealed. In this way, the generative model truly grasps and understands the principles governing data generation. Latent variables that are as independent as possible signify that the generative model possesses enhanced interpretability and clearer generation rules. Therefore, utilizing generative models to perform disentangling or ICA tasks is justified. 
    
        In recent years, VAEs have been applied to disentangling (\cite{higgins2017beta}; \cite{yingzhen2018disentangled}; \cite{burgess2018understanding}; \cite{kim2018disentangling}; \cite{chen2018isolating}; \cite{li2019disentangled}; \cite{locatello2019challenging}); \cite{klindt2020towards}; \cite{locatello2020weakly}), causal inference (\cite{pearl2019seven}; \cite{scholkopf2021toward}; \cite{yang2021causalvae}; \cite{lippe2022citris}; \cite{brehmer2022weakly};\cite{lippe2022citris}; \cite{ahuja2023interventional}; \cite{wendong2024causal}), and ICA(\cite{khemakhem2020variational};
        \cite{halva2021disentangling};  \cite{lachapelle2022disentanglement};
        \cite{wei2024half}; \cite{wei2025half}). These approaches have, to varying degrees, improved or altered the prior configurations of the original VAE (referred to as the vanilla VAE in this article). Gaussian processes (GPs) are probabilistic models capable of modeling temporal or spatial structures (\cite{seeger2004gaussian}; \cite{bishop2006pattern}). This property makes GPs particularly well suited for serving as priors for the latent variables with structured or correlated characteristics (\cite{casale2018gaussian}; \cite{tonekaboni2022decoupling}), including tasks such as data imputation (\cite{ramchandran2021longitudinal}) and trajectory inference (\cite{pearce2020gaussian}). By assigning separate GP priors to each dimension of the latent variables, it encourages each latent dimension to possess distinct structures (\cite{wei2024innovative}), thereby promoting disentanglement among them. The methods above for addressing different issues can all be classified under Gaussian Process Variational Autoencoders (GP-VAEs). However, GP-VAEs impose a significant computational burden when handling large datasets due to operations on an $n \times n$ covariance matrix (such as matrix inversion or eigen-decomposition), where $n$ is the number of data points. This results in a computational complexity of $\mathcal{O}(n^3)$ (\cite{quinonero2005unifying}; \cite{williams2006gaussian}). In other words, the computational time and resource usage associated with GPs increase significantly as the dataset size grows. 
        
        To identify alternatives to GPs, conducting an in-depth analysis of their underlying mechanisms is essential.  In GPs, specifying a kernel function enables the modeling of autocorrelations between data points based on their positions in the input domain, such as temporal or spatial dimensions. In other words, the kernel function defines the covariance structure between any two points within this domain. This covariance structure captures the functional characteristics that a GP can model, including smoothness, periodicity, and amplitude. Therefore, by assigning independent GP priors to each dimension of the latent variables to achieve the structuring or disentanglement of these latent variables, we are essentially modeling each dimension of the latent variables using different kernel functions. In other words, it is different kernel functions that are employed to model the distinct structures of each dimension of the latent variables. Based on the above analysis, alternatives to GPs should be selected from other methods that utilize kernel functions.
        
        Kernel regression is a non-parametric statistical method used to estimate the conditional expectation of a random variable. At the core of kernel regression lies the kernel function, which determines the weight assigned to each data point based on its distance from the target point  (\cite{rosenblat1956remarks}; \cite{nadaraya1964estimating}). The kernel function plays a pivotal role in shaping the estimation by controlling the smoothness and flexibility of the resulting regression curve. Similar to GPs, in kernel regression, the kernel functions directly affect how local or global the influence of data points is. Therefore, we aim to achieve effects analogous to those of GP-VAE by assigning distinct kernel regression functions to the priors of different dimensions of the VAE's latent variables. Specifically, each kernel regression function models the temporal or spatial structure associated with its respective latent variable dimension. This approach allows independent components to be differentiated based on these distinct latent dimensions, each modeled by a separate kernel regression function. This method is called the Structured Kernel Regression Variational Autoencoder (SKR-VAE). When separating a set of mixed data into \( N \) independent components, each with a sequence length of \( L \), the time complexity of GP-VAE for performing GP-related computations is \( \mathcal{O}(N\times L^3)\), whereas SKR-VAE has a time complexity of \(\mathcal{O}(N\times L^2)\). Consequently, SKR-VAE theoretically demonstrates a clear efficiency advantage, primarily because kernel regression does not involve matrix inversion operations. 
        
        Experiments are conducted to validate the effectiveness of the proposed method. The results demonstrate that the method significantly reduces GPU memory usage and greatly improves computational speed while maintaining ICA performance. This indicates that SKR-VAE has the potential to efficiently replace GP-VAE in solving ICA problems. The primary contributions of this study can be summarized as follows:
        
        (1). This study introduces the Structured Kernel Regression Variational Autoencoder (SKR-VAE) for the first time and validates its capability in solving ICA problems.
        
        (2). This study provides a detailed comparison of the computational efficiency between SKR-VAE and GP-VAE across various data scales, highlighting the advantages of SKR-VAE.

    \section{Structured Kernel Regression Variational Autoencoder}
        \subsection{VAE for ICA}
            Variational Autoencoders (VAEs) integrate the architecture of autoencoders with the principles of variational inference. From the perspective of variational inference, the objective of a VAE is to infer the posterior distribution \( p(\mathbf{Z}|\mathbf{X}) \) of the latent variables \( \mathbf{Z} \) given the observed data \(\mathbf{X} \). This inference process can be approximately equated to the inverse process in ICA. Assuming that the observed data \( \mathbf{X} \) is obtained from the latent variables \( \mathbf{Z} \) through a linear or nonlinear mapping \( \mathcal{F} \), i.e., \( \mathbf{X} = \mathcal{F}(\mathbf{Z})\), the encoder and decoder of the VAE form a complete separation \(\mathbf{Z} = \mathcal{F}^{-1}(\mathbf{X})\) and reconstruction  \( \widehat{\mathbf{X}} = \mathcal{F}(\mathbf{Z})\) loop, as illustrated in Figure 1. In this process, the choice of the prior \( p(\mathbf{Z}) \) for the latent variables is critically important as it directly affects the accuracy of the posterior \( p(\mathbf{Z}|\mathbf{X}) \). However, the vanilla VAE sets the prior \( p(\mathbf{Z}) \) to a standard normal distribution (\cite{kingma2013auto}; \cite{rezende2014stochastic}; \cite{kingma2019introduction}), which, although it facilitates easy sampling during the generative step of the VAE as a generative model, limits the inference capabilities of the VAE as a variational inference algorithm. Therefore, to obtain the posterior \( p(\mathbf{Z}|\mathbf{X}) \) of the independent components,  more reasonable priors need to be introduced, including GPs and the Structured Kernel Regression proposed in this study.

            In the variational inference algorithm, we approximate the posterior distribution \( p(\mathbf{Z}|\mathbf{X}) \) by introducing an auxiliary distribution \( q(\mathbf{Z}|\mathbf{X}) \), as shown below:
            \begin{equation}
            	D_{KL} (q(\mathbf{Z}|\mathbf{X}) \| p(\mathbf{Z}|\mathbf{X})) = \ln p(\mathbf{X}) + D_{KL} (q(\mathbf{Z}|\mathbf{X}) \| p(\mathbf{Z})) - \mathbb{E}_{q(\mathbf{Z}|\mathbf{X})} [\ln p(\mathbf{X}|\mathbf{Z})].
            \end{equation}
            Considering the non-negativity of KL divergence, the above equation can be transformed into the following inequality:
            \begin{equation}
            	\ln p(\mathbf{X}) \geq \mathbb{E}_{q(\mathbf{Z}|\mathbf{X})} [\ln p(\mathbf{X}|\mathbf{Z})] - D_{KL} (q(\mathbf{Z}|\mathbf{X}) \| p(\mathbf{Z})).
            \end{equation}
            The right-hand side of the inequality is known as the Evidence Lower Bound (ELBO). By maximizing the ELBO, the \( D_{KL} (q(\mathbf{Z}|\mathbf{X}) \| p(\mathbf{Z}|\mathbf{X})) \) term in Equation (1) can be minimized towards zero, allowing \( q(\mathbf{Z}|\mathbf{X}) \) to approximate the true posterior \( p(\mathbf{Z}|\mathbf{X}) \) as closely as possible. The ELBO consists of two main terms: the first term, \( \mathbb{E}_{q(\mathbf{Z}|\mathbf{X})} [\ln p(\mathbf{X}|\mathbf{Z})] \), indicates that the VAE aims to maximize the likelihood of the observed data by leveraging the inferred approximate posterior \( q(\mathbf{Z}|\mathbf{X}) \); the second term, \( -D_{KL} (q(\mathbf{Z}|\mathbf{X}) \| p(\mathbf{Z})) \), shows that the inferred \( q(\mathbf{Z}|\mathbf{X}) \) should be as close as possible to the prior \( p(\mathbf{Z}) \). Thus, the structured design of the prior \( p(\mathbf{Z}) \) can directly influence the inferred posterior \( q(\mathbf{Z}|\mathbf{X}) \) when maximizing the ELBO. In this study, we treat each dimension of the latent variable \( \mathbf{Z} \) as independent components. Therefore, under a logarithmic scale, the KL term in Equation (2) can be further expressed as:
            \begin{equation}
            	D_{KL} (q(\mathbf{Z}|\mathbf{X}) \| p(\mathbf{Z})) = \sum_{i=1}^N D_{KL} (q(\mathbf{Z}_i|\mathbf{X}) \| p(\mathbf{Z}_i)).
            \end{equation}
            \begin{figure}[ht]
            	\centering
            	\includegraphics[width=1\textwidth]{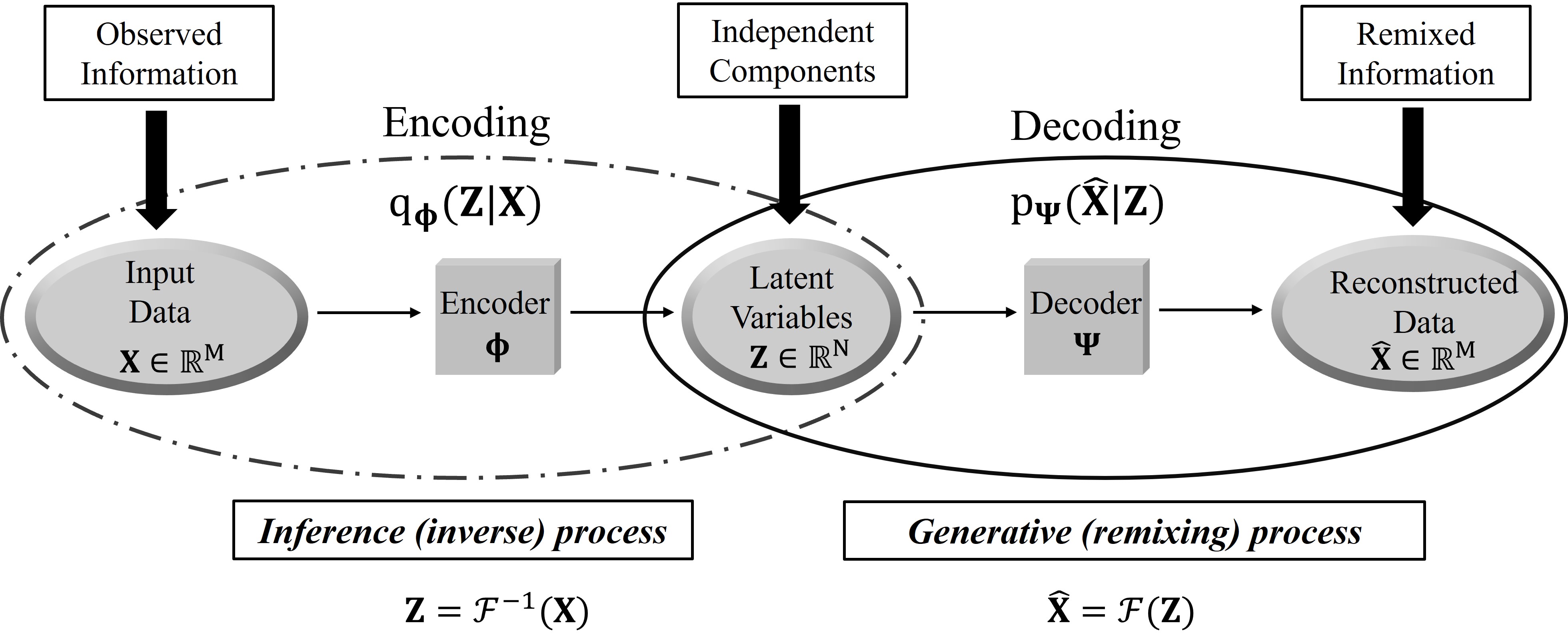}
            	\caption{VAE for the inference of independent components}
            	\label{VAE for the inference of independent components}
            \end{figure}
            
\subsection{Structured Kernel Regression}
Equation (3) assumes that each dimension of the latent variables is independent of the others. In this study, the independence among different dimensions of the latent variables arises from their distinct autocorrelations (whether temporal or spatial). In methods like GP-VAE, the distinct autocorrelation of each latent dimension is derived from being assigned a different GP. This can be represented as:
\begin{equation}
	\sum_{i=1}^N D_{KL} \left( q(\mathbf{Z}_i|\mathbf{X}) \| p(\mathbf{Z}_i) \right) = \sum_{i=1}^N D_{KL} \left( q(\mathbf{Z}_i|\mathbf{X}) \|\mathcal{GP}(\mathbf{Z}_i) \right).
\end{equation}
Correspondingly, in the SKR-VAE proposed in this study, different latent dimensions are modeled using different kernel regression functions. Thus, in SKR-VAE, the  kernel regression function (\(\mathcal{KRF}\)) replaces the role of each prior \( p(\mathbf{Z}_i) \) in the original formulation, as follows:
\begin{equation}
	\sum_{i=1}^N D_{KL} \left( q(\mathbf{Z}_i|\mathbf{X}) \| p(\mathbf{Z}_i) \right) = \sum_{i=1}^N D_{KL} \left( q(\mathbf{Z}_i|\mathbf{X}) \|\mathcal{KRF}(\mathbf{Z}_i) \right).
\end{equation}

For both Gaussian Process and kernel regression, their commonality lies in relying on kernel functions to define the correlation between data points, and using this correlation to construct the model's characteristics and structure. The Radial Basis Function (RBF) kernel, often used in GP-VAE (\cite{pearce2020gaussian}; \cite{tonekaboni2022decoupling}; \cite{wei2024half}), is also introduced in SKR-VAE, with the expression:
\begin{equation}
	k(\tau, \tau';\gamma) = \exp\left( -\frac{\|\tau - \tau'\|^2}{\gamma} \right),
\end{equation}
where \(\tau\) represents the index of the latent variable \(z\), and \(\|\tau - \tau'\|^2\) denotes the Euclidean distance between the indexes of \(z\) and \(z'\). The parameter \(\gamma\) controls the spread of the RBF kernel: a larger \(\gamma\) value means that the function value decays more slowly, resulting in a larger range of influence; conversely, a smaller \(\gamma\) value means rapid decay, limiting influence to a local region. Thus, the strength of influence between data points in the RBF kernel is determined by the parameter \(\gamma\). In short, the temporal or spatial structure of each latent dimension \(Z_i\) is modeled by its corresponding \(\gamma_i\).In SKR-VAE, the parameter \(\gamma\), along with the parameters of the encoder and decoder, is optimized through the objective function rather than being set in advance. The above process is integrated into the inference procedure of SKR-VAE, as shown in Figure 2.
	\begin{figure}[ht]
		\centering
		\includegraphics[width=1\textwidth]{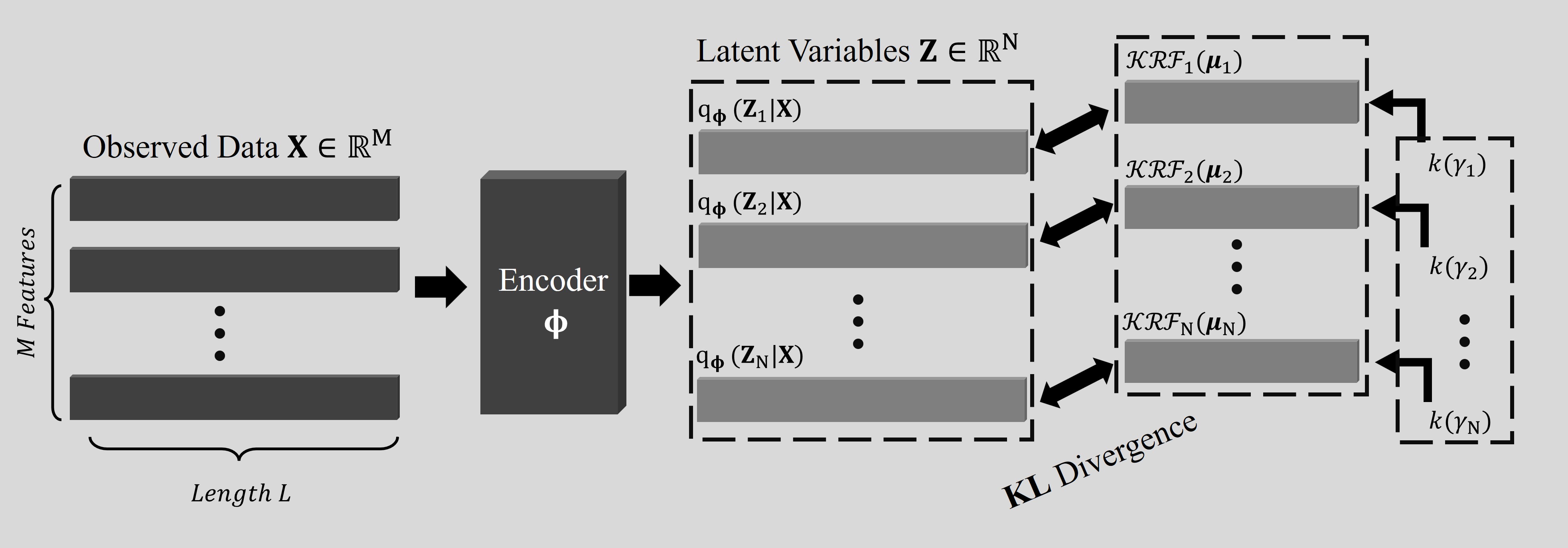}
		\caption{Inference process of SKR-VAE}
		\label{Inference process of SKR-VAE}
	\end{figure}

For each independent latent $\mathbf{Z}_i$ with dimension $\mathbb{R}^{1 \times L}$, the corresponding kernel regression function can be expressed as:
\begin{equation}
\boldsymbol{\mu}_{\mathcal{KRF}}^i = \frac{\sum_{j=1}^L k_i(\tau, \tau_j; \gamma_i) z_j^i}{\sum_{j=1}^L k_i(\tau, \tau_j; \gamma_i)}.
\end{equation}
In the term $D_{KL}(q(\mathbf{Z}_i|\mathbf{X})\|\mathcal{KRF}(\mathbf{Z}_i))$, $\mathcal{KRF}(\mathbf{Z}_i)$ essentially represents a probability distribution rather than a deterministic sequence. Thus, the above kernel regression function \(\mu_{\mathcal{KRF}}^i\) can be regarded as the mean of the distribution of the latent dimension, while the variance can be introduced as an additional hyperparameter (as a pseudo-variance) or directly adopted from the inferred variance of $q(\mathbf{Z}_i|\mathbf{X})$. The calculation of the KL term depends on the assumption of the $\mathcal{KRF}(\mathbf{Z}_i)$ distribution. In this study, this term is considered as a factorized Gaussian distribution, just like that of $q(\mathbf{Z}_i|\mathbf{X})$. Therefore, the KL term can be expressed analytically as:
\begin{equation} 
D_{KL}(q(\mathbf{Z}_i|\mathbf{X})\|\mathcal{KRF}(\mathbf{Z}_i))=\frac{1}{2}\left[ \ln\left(\frac{|\boldsymbol{\xi}_{\mathcal{KRF}}^i|}{|\boldsymbol{\xi}_q^i|}\right) - L + \text{tr}\left((\boldsymbol{\xi}_{\mathcal{KRF}}^i)^{-1} \boldsymbol{\xi}_q^i\right) + (\boldsymbol{\mu}_{\mathcal{KRF}}^i - \boldsymbol{\mu}_q^i)^\top (\boldsymbol{\xi}_{\mathcal{KRF}}^i)^{-1}(\boldsymbol{\mu}_{\mathcal{KRF}}^i - \boldsymbol{\mu}_q^i)\right],
\end{equation}
where $\boldsymbol{\xi}_{\mathcal{KRF}}^i$ and $\boldsymbol{\xi}_q^i$ represent the covariance matrices of $\mathcal{KRF}(\mathbf{Z}_i)$ and $q(\mathbf{Z}_i|\mathbf{X})$, respectively, and $\text{tr}(\cdot)$ denotes the trace of the matrix. In addition to the structured design mentioned above, an additional discriminator is also intruded to enforce the product of the marginal distributions of the latent variables to approximate their joint distribution (\cite{brakel2017learning}), i.e., \(p(Z_1, Z_2, \dots, Z_N) \approx p(Z_1) p(Z_2) \dots p(Z_N).\) This design provides effective assistance for SKR-VAE to solve the ICA problem.

\subsection{Optimization objectives of SKR-VAE}

To adapt to the characteristics of stochastic gradient descent in neural networks, maximizing the ELBO is transformed into minimizing its negative. By integrating the structured KL term, we obtain the loss function of SKR-VAE:
\begin{equation} 
L_{\text{SKR-VAE}}(\boldsymbol{\phi}, \boldsymbol{\Psi}, \boldsymbol{\xi}, \boldsymbol{\gamma}; \mathbf{X})) = \mathbb{E}_{q_{\boldsymbol{\phi}}(\mathbf{Z}|\mathbf{X})} [\ln p_{\boldsymbol{\Psi}}(\mathbf{X})|\mathbf{Z})] - \lambda \sum_{i=1}^N D_{KL} (q_{\boldsymbol{\phi}}(\mathbf{Z}_i|\mathbf{X})) \| \text{KRF}_{\gamma_i}(\mathbf{Z}_i)),
\end{equation}
where \(\boldsymbol{\phi}\) and \(\boldsymbol{\Psi}\) represent the parameters of the encoder and decoder, respectively. In addition to these parameters, the objective function also infers the kernel function parameters \(\boldsymbol{\gamma} = (\gamma_1, \gamma_2, \dots, \gamma_N)\). Each independent component posterior \(q_{\boldsymbol{\phi}}(\mathbf{Z}_i|\mathbf{X})\) is assigned only one variance \(\xi_i\), which means that the variances of an individual component posterior (which is a factorized Gaussian distribution) are shared. This design reduces model complexity and decreases the number of parameters that need to be inferred (\cite{wei2024half}). Similar to the kernel function parameters \(\boldsymbol{\gamma}\), \(\boldsymbol{\xi} = (\xi_1, \xi_2, \dots, \xi_N)\) is also optimized based on the loss function \(L_{\text{SKR-VAE}}\). Furthermore, \(\lambda\) serves as a hyperparameter that balances the weight of the KL divergence term and the reconstruction term. The process of SKR-VAE solving the ICA problem is essentially an model optimization according to the loss function \(L_{\text{SKR-VAE}}\).

\section{Experiments}
\subsection{Synthetic Signals}
We generate three synthetic source signals, each containing 10{,}000 samples, to serve as independent sources/independent components. The three sources are obtained by applying band-pass filters with increasing center frequency. Consequently, the lowest-frequency source exhibits the strongest sample-to-sample autocorrelation and is the smoothest, whereas the highest-frequency source has the weakest autocorrelation and varies the fastest. The three sources are thus distinguishable by their autocorrelation profiles. Interpreting the sample index as time, this autocorrelation corresponds to temporal structure; interpreting it as spatial position, it corresponds to spatial structure.

\begin{figure}[ht]
	\centering
	\includegraphics[width=1\textwidth]{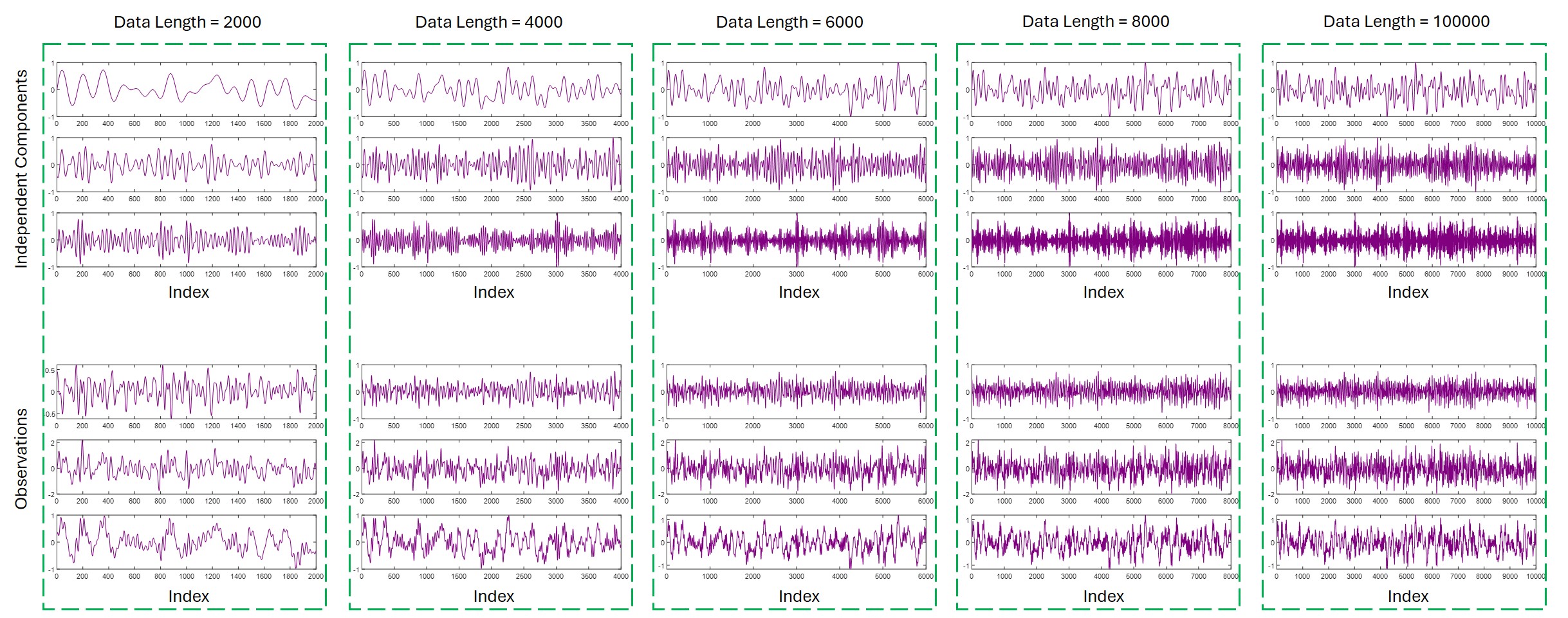}
	\caption{Independent components and mixed observations at different data lengths}
	\label{Independent components and mixed observations at different data lengths}
\end{figure}

\subsection{Experimental Setup}
To examine how the number of samples per signal affects GP-VAE and SKR-VAE, we construct a sequence of scenarios by truncating the three sources to different lengths: 2{,}000, 4{,}000, 6{,}000, 8{,}000, and the full 10{,}000 samples. For each length, we form the corresponding mixed observations. The ground-truth sources and their mixtures for all lengths are shown in Figure~3.

In addition to GP-VAE, we include three baselines for comparison: the vanilla VAE, $\beta$-VAE ($\beta\!=\!0.5$), and $\beta$-VAE ($\beta\!=\!2$). Performance is evaluated using the \emph{max correlation} criterion (following \cite{brakel2017learning}): for each recovered component we compute its maximum correlation with any ground-truth source, and report the average across components. Correlations close to $1$ indicate an excellent fit to the ground truth. Because ICA/BSS recovery is indeterminate up to component-wise scaling, correlation-based evaluation avoids the need for additional normalization.

\subsection{ICA (BSS) Performance}
Figure~4 and Table~1 reports ICA performance across sequence lengths. SKR-VAE and GP-VAE consistently outperform the other baselines. Figure~5 shows the corresponding wall-clock time. GP-VAE incurs a substantially higher time cost than all other methods, and its cost accelerates as the number of samples increases, reflecting the well-known scalability challenge of GP-based approaches. 

\begin{figure}[!htbp]
	\centering
	\includegraphics[width=1\textwidth]{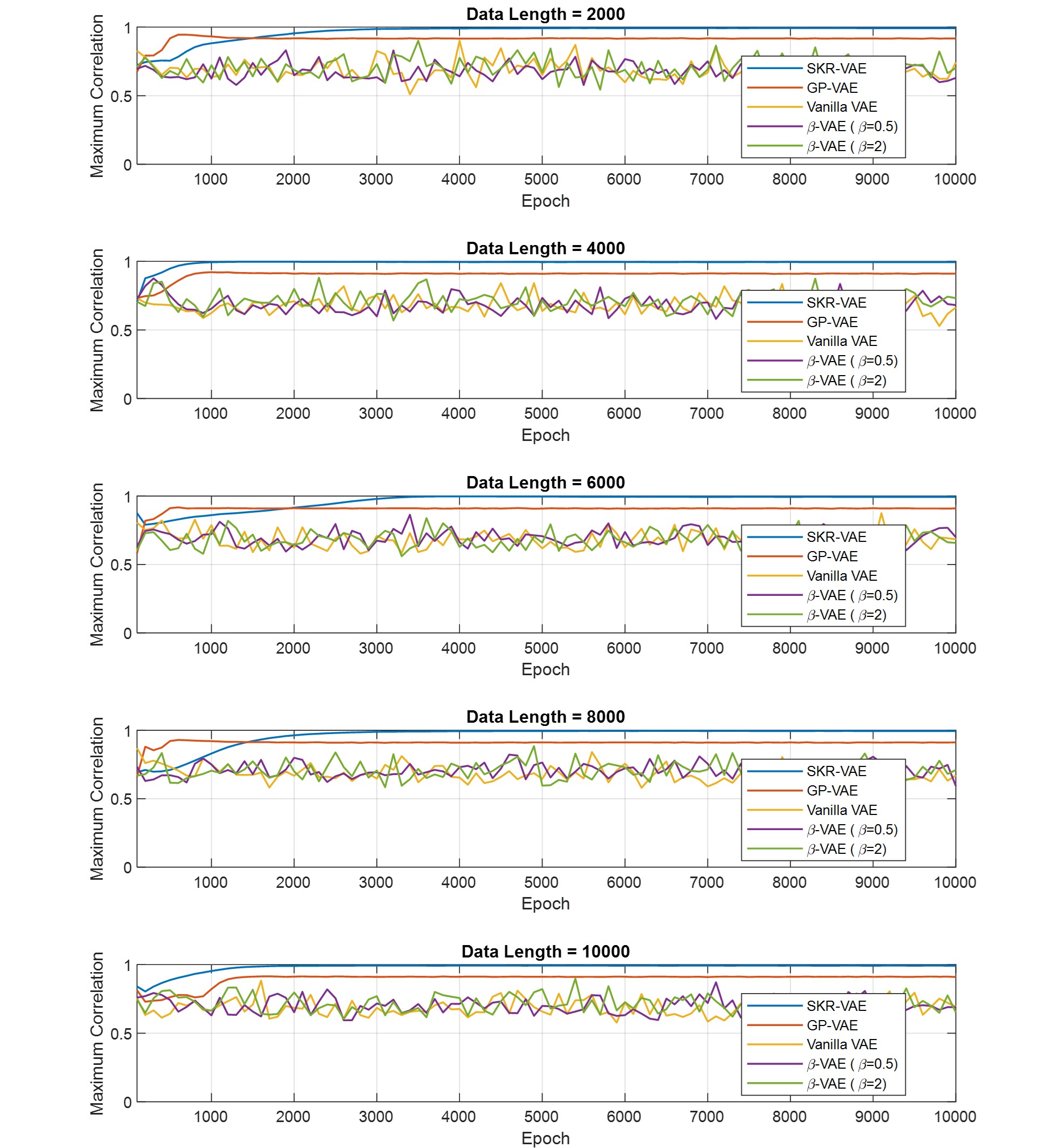}
	\caption{Convergence of different methods}
	\label{Convergence of different methods}
\end{figure}

\begin{table}[htbp]
	\centering
	\caption{Max correlations}
	\begin{tabular}{cccccc}
		\toprule
		\textbf{Data length} & \textbf{skrVAE} & \textbf{gpVAE} & \textbf{VAE} & \textbf{betaVAE(0.5)} & \textbf{betaVAE(2)} \\
		\midrule
		2000  & \textbf{0.9921} & 0.9167 & 0.7435 & 0.6299 & 0.7026 \\
		4000  & \textbf{0.9940} & 0.9113 & 0.6636 & 0.6811 & 0.7322 \\
		6000  & \textbf{0.9938} & 0.9100 & 0.6838 & 0.6986 & 0.6578 \\
		8000  & \textbf{0.9950} & 0.9116 & 0.6704 & 0.5927 & 0.7102 \\
		10000 & \textbf{0.9924} & 0.9115 & 0.6941 & 0.6899 & 0.6522 \\
		\bottomrule
	\end{tabular}
	\label{tab:comparison_data_length}
\end{table}

While SKR-VAE is also more expensive than the non-GP baselines, its time cost remains within a practical range. Taken together with the accuracy in Figure~4, SKR-VAE provides a superior trade-off to GP-VAE. In particular, GP-VAE is approximately two orders of magnitude slower than SKR-VAE in our settings. All methods are executed under identical hardware conditions.

\begin{figure}[ht]
	\centering
	\includegraphics[width=1\textwidth]{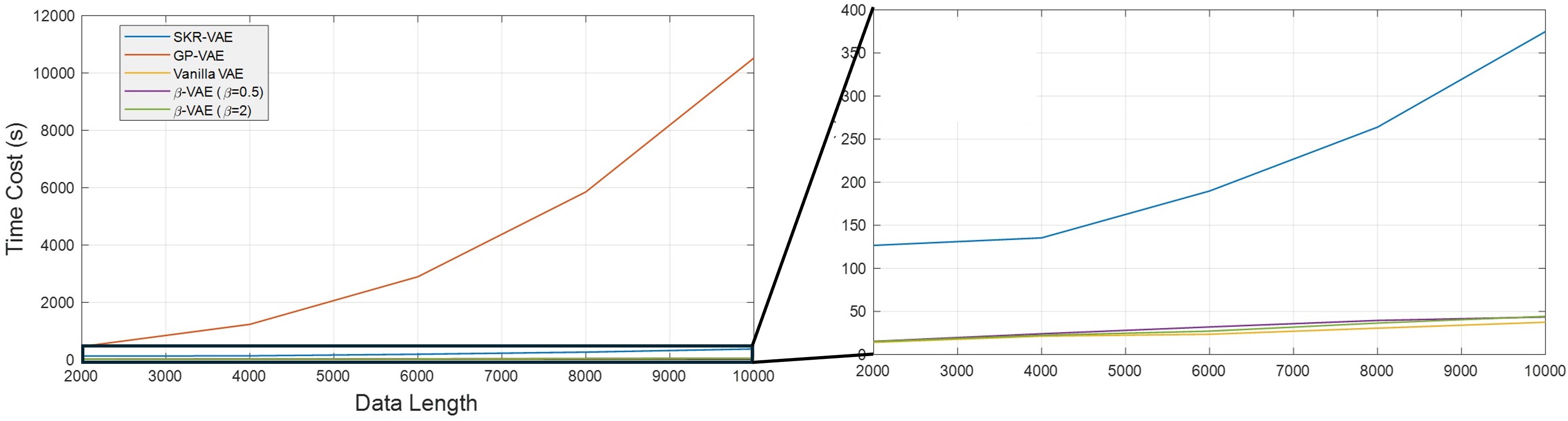}
	\caption{Time costs}
	\label{Time costs}
\end{figure}

\section{Discussion and Conclusion}
Structuring the latent space of a VAE facilitates the decomposition of mixed information into structured and mutually independent sources (latent variables). In the context of variational inference, the choice of prior for the latent variables plays a pivotal role. SKR-VAE and GP-VAE share a similar design philosophy: both aim to separate one-dimensional sequences according to differences in their autocorrelation patterns (temporal or spatial), which are modeled by kernel functions with adaptive parameters. In GP-VAE, the kernel function is embedded into the modeling of intra-dimensional correlations via the covariance (kernel) matrix in the marginal likelihood, which requires matrix inversion operations. In contrast, SKR-VAE achieves a similar goal through kernel regression, avoiding explicit matrix inversion.

Although both approaches rely on kernel functions to model the structural properties of latent variables, GP-VAE suffers from a steep increase in computational cost as the dataset grows, whereas SKR-VAE maintains substantially higher computational efficiency. This indicates that, when kernel functions are explicitly required, adopting a more computationally efficient integration strategy can be advantageous. Based on the above analysis, SKR-VAE can serve as an effective and efficient alternative to GP-VAE. 

Future research will further validate the capabilities of SKR-VAE in broader contexts, including nonlinear ICA, disentanglement, interpretable generative modeling, and causal inference.

\FloatBarrier
\bibliography{ref.bib}

\end{document}